%% file: main.tex
\documentclass{article}
\PassOptionsToPackage{numbers,compress}{natbib}
\usepackage[preprint]{neurips_2026}
\usepackage[utf8]{inputenc}
\usepackage[T1]{fontenc}
\usepackage{hyperref,url,booktabs,array,tabularx,amsmath,amsfonts,microtype,graphicx,xcolor}
\usepackage{placeins}
\definecolor{guardteal}{HTML}{007D88}
\hypersetup{colorlinks=true,linkcolor=guardteal,citecolor=guardteal,urlcolor=guardteal}
\newcommand{\mg}{MedGuard}
\newcommand{\pp}{\,\mathrm{pp}}
\newcolumntype{L}{>{\raggedright\arraybackslash}X}

\title{MedToolFault and MedGuard: evaluating and improving clinical agent reliability}
\input{authors.tex}
\input{figure_pages.tex}
\begin{document}
\maketitle

\begin{abstract}
Clinical agents must retrieve patient-specific evidence and correctly update
records despite unreliable tool responses. We introduce MedToolFault, a
fault-injection benchmark linking six tool-failure classes to answer correctness, required
state changes and forbidden mutations in public clinical-agent tasks. We also
introduce MedGuard, a tool-layer method that verifies read
evidence and attributes ambiguous writes through public interfaces. Across
127,861 formal episode evaluations, we assess fault recovery, mechanism
controls, task expansion, constrained tool budgets and external question
answering. Across nine model configurations, unguarded agents fail in
32.2--67.8\% of fault-injected evaluations on tasks each configuration solves
without faults. MedGuard improves paired safe success by 13.7--46.6 percentage
points on these tasks.

\end{abstract}

\section{Introduction}

Medical AI systems are moving beyond single-turn answers towards diagnostic
dialogue and longitudinal disease management~\citep{tu2025,lievin2026}.
Tool-using agents extend this progression by gathering evidence from external
resources and acting within electronic health record (EHR)
workflows~\citep{zhao2026,jiang2025}. Their clinical reasoning is consequently
coupled to the information acquired during interaction, as sequential
diagnostic evaluations demonstrate~\citep{chen2026,schmidgall2026}.
Recent clinical-agent studies further distinguish diagnostic accuracy from
decision-time reliability and document writes that proceed despite
patient-identity inconsistencies~\citep{zhang2026,klang2026}. For agents that
can change a patient record, these findings motivate evaluating the integrity
of acquired evidence and completed actions alongside the final
answer~\citep{jiang2025,klang2026,yao2025}.

Existing clinical benchmarks address different parts of this workflow.
ClinDiag-Benchmark~\citep{chen2026} and AgentClinic~\citep{schmidgall2026}
assess sequential information gathering and diagnosis, while
FHIR-AgentBench~\citep{lee2026} measures question answering over
interoperable records. These evaluations
characterize clinical reasoning and retrieval, but their diagnostic and
answer-level endpoints do not establish whether a requested record update
was correctly committed. MedAgentBench~\citep{jiang2025} introduces both retrieval and action
tasks in a virtual EHR; its original action-evaluation protocol checks POST
payloads rather than executing those writes~\citep{bellibatlu2026}.
A valid payload alone cannot distinguish a completed action from a failed
attempt, or reveal whether retrying an uncertain outcome creates a duplicate.
The unresolved issue is therefore execution under unreliable feedback, beyond
the ability to formulate an appropriate clinical answer or action.

General-purpose agent benchmarks provide complementary evaluation methods.
$\tau$-bench~\citep{yao2025} checks final database states and repeated-run
consistency; ToolBench-X~\citep{tian2026} measures recovery from tool hazards
against canonical final answers. ReliabilityBench~\citep{gupta2026} combines
API faults with state-based checks in scheduling, travel, customer support and
e-commerce. AgentChaos~\citep{tan2026} targets model-API failures, whereas
IdempotencyBench~\citep{swamy2026} and counterfactual recovery
studies~\citep{sun2026} examine duplicate effects and ambiguous action outcomes. These studies
establish that fault recovery and state verification matter, but their reported
evaluations do not establish whether recovery preserves clinical evidence
requirements together with patient-record integrity. In an EHR workflow, the
returned resource must refer to the intended patient and satisfy the requested
code, status and temporal constraints; any subsequent write must also achieve
the required change without introducing forbidden mutations.

Clinical reliability studies sharpen this distinction. Patient-identity
experiments test whether agents withhold writes when records are
inconsistent~\citep{klang2026}, and repeated-run studies reveal variation in
emitted clinical orders that a task-level verdict can conceal~\citep{bellibatlu2026}.
Decision-time confidence methods address when diagnostic outputs should be
trusted or deferred~\citep{zhang2026}. Our question instead concerns recovery:
can an agent complete a patient-specific task after a tool failure while
preserving both evidence validity and the required record state? Withholding
a write may prevent a forbidden change yet leave a required action undone;
repeating it may restore completion or create a duplicate, depending on whether
the earlier attempt committed. A clinically useful recovery evaluation must
therefore distinguish successful completion, omission and unsafe mutation,
and test how the same recovery mechanism responds to faults in both reads and
writes. This distinction is necessary to determine whether a safeguard
restores the workflow or merely shifts the type of failure.

Here we introduce MedToolFault, a fault-injection benchmark that links
controlled tool disturbances to patient-specific evidence and record-state
contracts, and \mg{}, a tool-layer verification and recovery method.
MedToolFault builds on public clinical-agent tasks
and reference answers, instantiates six operational failure classes, and
independently evaluates answer correctness, required changes and forbidden
mutations. This design makes the consequences of a fault observable beyond
the agent's completion claim. MedGuard addresses the corresponding execution
problem by verifying read evidence and attributing ambiguous write outcomes
through documented public interfaces before selecting a recovery action
(Fig.~\ref{fig:overview}a--f).
The benchmark supplies the outcome checks; the method operates on publicly
accessible evidence, without access to the evaluator's hidden contracts.

We test this benchmark--method combination using instructed and evidence-only
controls, component ablations and a shared tool-call ceiling to separate
verification from additional guidance or tool access. Task expansion and
external question answering assess scope, while delayed visibility and
concurrent writes test the assumptions needed for write attribution.
Multiple models evaluate the framework across agent implementations, with
detailed sensitivity analyses in Supplementary Analysis~S7.

\section{Results}

\subsection{MedToolFault connects tool failures to clinical task and state contracts}

MedToolFault combines patient-specific tasks, controlled disturbances at the
public tool interface, and explicit checks of both answers and record state.
It builds on public clinical-agent tasks and patient records. Its six failure classes cover
interface drift, transient service errors, stale or incomplete evidence,
patient mismatch, provenance conflict and ambiguous write outcomes
(Fig.~\ref{fig:overview}a--f). Committed and uncommitted timeouts are two
variants of the last class: the visible error
is identical, but the underlying state calls for different recovery actions.

The clinical specificity lies in the contracts connecting these disturbances
to patient-bound evidence and actions. A retrieved observation must belong
to the requested patient and satisfy the task's code, status and time
requirements; a completed write must create the required resource without
forbidden changes to the record. This design distinguishes an apparently
valid response from valid evidence, and a completion claim from a completed
action. An independent evaluator checks those contracts after execution.
Neither the agent nor MedGuard receives the hidden contract or reference
answer. Fault eligibility is assigned before execution, and untriggered
assignments remain in the denominator, preventing recovery rates from being
conditioned only on trajectories that reach the relevant tool call.

The MedAgentBench-derived substrate contains 300 unique tasks over ten source templates and
100 patient groups after removing 100 exact duplicates from a 400-record
source inventory. Patient-level partitioning yields 45 development tasks from
12 groups and 255 test tasks from 88 groups; 60 test tasks require writes.
Each configuration in the fault-recovery evaluation uses five policies on 255 clean and
806 eligible fault assignments, producing 5,305 episodes per configuration.
The nine configurations represent eight checkpoints from five model families;
Qwen3-8B is evaluated with thinking enabled and disabled.

Follow-ups test Qwen3.5-9B, Ministral-3-8B, Llama-3.1-8B and GPT-OSS-20B
through mechanism controls, expanded tasks, a shared raw-call ceiling and
external original-answer questions. Together with targeted adverse-action
checks, the study contains 127,861 formal evaluations, not independent
patients (Supplementary Table~S1). Development,
interrupted and superseded runs are excluded; pooled sampling summaries are
not counted again. Supplementary Methods~S6 documents the complete evidence
lineage and retained run inventory.

We define \emph{safe success} as a correct answer, correct required final-state
changes, and no forbidden mutation. \emph{Unsafe state} is a separate contract
violation endpoint, not a clinical injury rate. Omitting a required action
causes failure but is not, by itself, a forbidden mutation. The conditional
recovery endpoint uses the tasks solved by the same configuration's
vanilla-clean run. Whole-task effects are also reported because the
clean-capable subsets vary markedly across models. Patient-clustered intervals
describe uncertainty; they do not constitute multiplicity-adjusted tests of
all displayed comparisons.

\subsection{MedGuard verifies read evidence and attributes ambiguous writes}

MedGuard intercepts tool calls between the agent and the public tools.
Metadata-guided request repair addresses documented interface drift
(Fig.~\ref{fig:overview}a), and bounded retries address transient read failures
(Fig.~\ref{fig:overview}b). For stale or incomplete evidence, the read pathway
retrieves authoritative observations and follows trusted pagination
(Fig.~\ref{fig:overview}c). Patient-reference checks identify mismatched
evidence (Fig.~\ref{fig:overview}d), while source and finalized-status checks
resolve conflicting responses (Fig.~\ref{fig:overview}e). These operations
preserve requested ordering and count semantics and use the same public
interfaces available to the baseline agent.

For ambiguous write outcomes, MedGuard compares equivalent records before and
after a patient-linked creation to determine whether to recognize a commit
or repeat the write (Fig.~\ref{fig:overview}f). The matching criteria,
decision rules and operating assumptions are specified in Methods.

\OverviewFigure
\FloatBarrier

\subsection{Safe success, unsafe states and duplicate writes under tool faults}

On tasks that each configuration solves without faults, unguarded execution
fails in 32.2--67.8\% of assigned fault evaluations. These are raw
assignment-weighted failure rates from the conditional subset (Source Data),
whereas the paired recovery gains below weight each eligible task equally.

All nine configurations in the fault-recovery evaluation exhibit a positive conditional paired
MedGuard-minus-vanilla effect, ranging from $13.71\pp$ to $46.64\pp$, with each
95\% interval above zero (Fig.~\ref{fig:primary}b; Table~\ref{tab:primary}).
For Qwen3-8B thinking, raw vanilla
clean success is 87.8\%, raw vanilla fault success is 44.8\%, and raw guarded
fault success is 88.6\%. Its conditional paired gain is $46.64\pp$
[$42.00,51.02$]. Qwen3.5 improves from 51.2\% to 75.9\% and Ministral from
47.5\% to 75.6\% in the raw fault-assignment proportions. These proportions
weight assignments equally; they need not equal task-weighted paired effects.

The conditional endpoint separates recovery on solvable tasks from underlying
task competence. Whole-task effects and eligibility counts are retained
alongside it (Fig.~\ref{fig:primary}b; Table~\ref{tab:primary}); the all-task
interval for Qwen2.5-14B crosses zero, so the conditional result does not imply
all-task superiority for every configuration. The fault-wise map locates
recovery by disturbance type (Fig.~\ref{fig:primary}c), without estimating
real-world fault prevalence. Supplementary Analysis~S7 and Fig.~S1 provide
the detailed model- and template-stratified interpretation.

\begin{table}[!htbp]
\caption{Safe-success rates and paired recovery gains under tool faults. Rates are percentages. V, vanilla;
MG, MedGuard; $n_c$, vanilla-clean-capable tasks. Conditional gains are
task-weighted percentage-point differences with 95\% patient-cluster intervals,
not differences between the displayed all-assignment rates.}
\label{tab:primary}
\centering\small\setlength{\tabcolsep}{3pt}
\begin{tabular}{@{}lrrrrrr@{}}
\toprule
Configuration & V clean & V fault & MG fault & $n_c$ & Gain & 95\% interval\\
\midrule
\input{tables/primary.tex} 
\bottomrule
\end{tabular}
\end{table}

MedToolFault reports record integrity separately from successful completion
(Fig.~\ref{fig:primary}d). Among 202 action-required fault assignments per
configuration, unsafe-state counts fall from 36 to 13 for Qwen3.5,
73 to 48 for Ministral, and 67 to 5 for Gemma. They rise from 125 to 174 for
Llama and from 22 to 71 for Phi; Qwen2.5-14B and Qwen3 thinking also show
increases. These counts limit the safety interpretation of the recovery
result: improving safe-success frequency is not sufficient to establish an
overall reduction in unsafe-state frequency.

The ambiguity-specific duplication test supports a narrower conclusion.
Across all nine configurations, guarded execution produces no duplicates in
the 60 committed-timeout assignments per configuration
(Fig.~\ref{fig:primary}e). For Qwen3 thinking, ordinary retry produces
51 duplicates, compared with five for vanilla and
zero for \mg{}. The matched uncommitted-timeout assignments have no duplicates
for any of these three policies. Thus treating a lost response as an
unexecuted operation can cause over-execution. However, no duplicate does not
mean a correct required action occurred; Llama and Phi already have zero
vanilla duplicates in this test despite poor overall execution.

\RecoveryFigure
\FloatBarrier

\subsection{Instructed controls and component ablations}

The matched-control experiment compares instructed and evidence-only execution
with MedGuard. Informed agents receive explicit recovery instructions and the
same public interfaces; evidence-only agents receive verification evidence
without automatic selection or commit attribution. Raw fault success for
vanilla, informed, evidence-only and \mg{} is 53.8, 51.2, 56.7 and 71.8\%
for Qwen3.5; 49.0, 59.9, 57.6 and 79.4\% for Ministral; and 70.0, 69.4,
71.0 and 92.2\% for GPT-OSS (Fig.~\ref{fig:mechanisms}a).

The paired advantage over informed execution is $14.25\pp$ [$10.52,17.99$]
for Qwen3.5, $18.86\pp$ [$15.19,22.74$] for Ministral and $17.53\pp$
[$14.79,20.38$] for GPT-OSS. The corresponding Llama interval crosses zero,
so this control supports an advantage in three of four tested implementations,
not all four (Fig.~\ref{fig:mechanisms}b). The complete policy-specific rates,
paired estimates and adverse state outcomes are retained in the figures,
Source Data and Supplementary Analysis~S7.

Deleting read verification reduces paired success by 6.53, 15.63 and
9.81 percentage points in Qwen3.5, Ministral and GPT-OSS, respectively;
deleting write attribution reduces it by 4.79, 6.22 and 7.38 points.
Each of these deletion intervals is below zero. Read deletion also reduces
Llama success, whereas its write-deletion interval crosses zero
(Fig.~\ref{fig:mechanisms}c). All four generic-retry/interface-repair deletion
intervals cross zero. These ablations support contributions from read
verification and write attribution to the observed recovery gains.

\MechanismsFigure
\FloatBarrier

\subsection{Recovery under a shared call ceiling and consistency stress}

The shared-budget experiment fixes the maximum at 12 raw public tool calls,
including automatic verification and retry calls. On the same 50 tasks from
40 patient groups, guarded paired gains over vanilla are $16.12\pp$
[$8.01,24.46$] for Qwen3.5, $21.03\pp$ [$13.33,29.50$] for Ministral and
$14.48\pp$ [$8.49,20.54$] for GPT-OSS (Fig.~\ref{fig:mechanisms}d).
Llama's $3.40\pp$ interval [$-3.47,10.44$] crosses zero. These findings
exclude a larger nominal raw-call allowance as the sole explanation for the
three positive effects, but do not equate realized calls, token use or runtime.

Repeated and compound faults retain positive effects for the same three
models (Fig.~\ref{fig:stress}a). Sampling and output-budget checks assess
sensitivity to execution settings (Fig.~\ref{fig:stress}b; Supplementary
Analysis~S7). These checks reuse the same tasks across execution settings,
measuring recovery stability within a fixed patient population.

Fault placement matters. A standard transient fault can be consumed by a
guard preflight operation. When the fault is instead aligned to the agent's
target operation, GPT-OSS has a paired effect of $-2.75\pp$
[$-4.98,-0.76$], despite its positive matched-control effect
(Fig.~\ref{fig:stress}d). This contrast shows that recovery depends on where
a fault occurs relative to verification.

The consistency-boundary experiments test delayed visibility and concurrent writes. All four guarded
models achieve 0/60 successes when a committed write is temporarily hidden
from the verification read. Under a controlled concurrent writer, GPT-OSS
improves from 3/120 to 60/120 successes, but still fails half the assignments.
Its 58/120 guarded unsafe states include external-writer contributions and
cannot all be attributed to the agent. The wrapper therefore does not solve
general transaction attribution under concurrency or delayed visibility.

\StressFigure
\FloatBarrier

\subsection{Recovery and state outcomes on expanded clinical tasks}

The 470 test variants in task expansion probe decision branches, preexisting
equivalent/different/cancelled records, and rule-based actions over external
data. They are semi-synthetic operational contracts, not newly validated
treatment recommendations. On decision branches, paired gains are
$24.87\pp$ for Qwen3.5, $22.85\pp$ for Ministral and $26.56\pp$ for
GPT-OSS, with intervals above zero; Llama remains near zero
(Fig.~\ref{fig:transfer}a). Yet Ministral's action-required unsafe states
increase from 172/504 to 307/504. Improvement in the aggregate success
endpoint can therefore coexist with deterioration in a safety-relevant stratum.
The disjoint action/no-action analysis shows that this aggregate improvement
does not reflect improved required-action execution (Supplementary Fig.~S3).

On external-source actions, GPT-OSS
improves from 519/732 to 726/732 fault successes and from 62/435 to 0/435
unsafe action-required states. Qwen3.5 instead falls from 481/732 to 297/732
successes, with a paired effect of $-25.46\pp$ [$-28.67,-22.10$], even as
its unsafe-state count falls from 12/435 to zero. Its clean success also
falls from 125/138 to 69/138. Reporting only unsafe-state reduction would
misrepresent this loss of task completion.

A post-hoc trace inspection identifies a concrete implementation limit in the
Qwen contrast: detailed query-parameter errors are replaced by generic
verification failures, followed by repeated invalid requests and missing
writes (Fig.~\ref{fig:transfer}c--e). All 56 newly lost clean and 234 newly
lost fault pairs share these observable tags. Supplementary Methods~S5 and
Analysis~S7 retain the pair counts and code-path interpretation. No
error-preserving replacement guard was evaluated, so this diagnosis does
not establish the efficacy of a proposed repair.

\TransferFigure
\FloatBarrier

\subsection{Answer accuracy on external read-only questions}

External question answering evaluates 100 original FHIR-AgentBench questions covering 45 patients
and 25 templates on MIMIC-IV-FHIR Demo data. The four models use the same
questions, five read-side faults and three policies. This external read-only
test retains original reference answers and differs from the rule-based
external-action tasks above. It does not evaluate write safety or represent
an independent hospital cohort.

Across 500 fault assignments per policy, Qwen3.5 answers 89 correctly with
vanilla, 108 with informed execution and 130 with \mg{}; GPT-OSS answers
133, 175 and 187, respectively (Fig.~\ref{fig:external}a). Their patient-equal
paired gains over vanilla are $8.59\pp$ [$3.85,13.85$] and $9.93\pp$
[$3.56,16.00$]. Ministral's effect is $-1.93\pp$ [$-5.33,1.26$] and
Llama's is $0.81\pp$ [$-0.15,1.93$]. All four guarded-minus-informed
intervals cross zero (Fig.~\ref{fig:external}b). We therefore do not establish
an external advantage over the instructed baseline; these intervals
also do not establish equivalence.

The all-question external endpoint includes all assigned questions rather than
small clean-capable subsets. Context and turn limits remain observed outcomes,
not discarded infrastructure records (Fig.~\ref{fig:external}d). Detailed
model-specific workflow markers and low-clean-competence sensitivity are
reported in Supplementary Fig.~S2 and Analysis~S7.

\ExternalFigure
\FloatBarrier

\subsection{Accumulated token use and raw tool calls}

In the matched-control experiment, mean accumulated token use decreases by 36.9\% for
Qwen3.5, 31.7\% for Ministral, 17.7\% for Llama and 30.6\% for GPT-OSS
under guarded execution, while mean raw tool calls increase for all four
(Fig.~\ref{fig:resources}). The same qualitative separation holds under the
shared 12-call ceiling. In that experiment, Qwen's token mean falls from 8,474 to 4,383, whereas raw calls
rise from 2.675 to 3.256. Tokens count accumulated model input and output,
including repeated conversation history, rather than distinct new words.

For external question answering, token use increases by 8.1\%
for Qwen, 6.0\% for Ministral and 0.4\% for Llama, and decreases by 10.9\%
for GPT-OSS. Raw-call means increase in all four. Thus an efficiency claim
must specify the workload and accounting unit. Neither fewer tokens nor a
shared call ceiling establishes lower end-to-end latency, energy consumption,
or deployment cost; these endpoints were not standardized in this study.

\ResourcesFigure
\FloatBarrier

\section{Discussion}

MedToolFault and MedGuard address two parts of clinical-agent
reliability: defining what should remain correct under tool failures, and
providing an execution-layer mechanism for recovering that correctness.
The benchmark connects controlled interface disturbances to patient-specific
evidence and record-state contracts. The method uses public verification
operations to check observations and distinguish ambiguous write outcomes.
Separate outcome assessment compares recovery policies against the same
clinical task and record-state contracts.

The nine-configuration evaluation shows that verification improves recovery on tasks
solvable without faults. Instructed controls, raw-evidence controls and
component ablations support contributions from read verification and write
attribution. Gains under a shared 12-call ceiling in three of
four follow-up implementations further show that a larger nominal tool
allowance is not the sole explanation. Together, these comparisons link
verification operations to measurable changes in task completion and
record-state integrity.

The clinically relevant lesson is the separation of correct language, completed
actions and record integrity. An omitted order and a duplicated order may
both fail a task, but they are not the same event. Likewise, a lower unsafe-state
rate can result from fewer actions being completed rather than more reliable
execution. In the Qwen external-action contrast,
unsafe states disappear while task success deteriorates. A medical-agent
benchmark that reports only answer accuracy or only prohibited mutations
would obscure this trade-off. Our use of state contracts makes the distinction
inspectable, but the contracts remain operational definitions, not a validated
scale of patient harm.

The results also motivate treating recovery evidence as part of the interface
contract. The two timeout worlds are indistinguishable from the response
alone; public state observations provide useful information only under
assumptions about visibility and attribution. Delayed writes and concurrent
writers violate those assumptions. Similarly, replacing a detailed error with
a generic verification message may remove the information an agent needs to
repair its request. This creates a design tension: preventing an
unchecked result from being presented as verified should not require hiding
actionable diagnostics. Error-preserving verification responses, explicit
unresolved-outcome types, and transaction-level commit attribution are
motivated follow-up directions, not solutions validated by the present data.

MedToolFault connects tool-fault testing and end-state evaluation to
patient-specific evidence integrity and write-state correctness. Its clinical
scope is defined by explicit patient, evidence and action contracts: these
contracts determine whether a recovered observation is usable and whether an
action has produced the intended record change. Coupling this evaluation with
MedGuard makes recovery mechanisms, their components and their operating
assumptions directly testable. The resulting framework exposes both recovered
tasks and unresolved failure modes, providing an inspectable basis for
subsequent recovery-method comparisons.

Several limitations constrain generalization. The public task substrates cover
limited templates and patient groups, and source-benchmark exposure during
model pretraining cannot be excluded. Repeated policies and faults expand
experimental coverage without expanding the independent patient population.
Some task-expansion rules are semi-synthetic and have not undergone new
clinical adjudication. Strict resource contracts may reject clinically
equivalent alternative encodings. Evaluator limitations themselves deserve
attention, as highlighted by prior clinical world-feedback audits~\citep{mantravadi2026}.
The proxy's authoritative endpoint and declared consistency assumptions are
controlled affordances, not a description of every real EHR installation.

Model-level effects vary with the agent's baseline competence and execution
behavior (Supplementary Analysis~S7). The model cohorts and execution
protocols also differ across stages.
Gemma results from the nine-configuration evaluation are retained, whereas GPT-OSS enters the
follow-ups. External questions are sampled from a previously unattempted patient
remainder after infrastructure difficulties interrupted an earlier evaluation;
this sampling frame does not cover the full external benchmark. Model-specific
serialization adaptations make within-model policy comparisons more defensible
than an unqualified cross-model leaderboard. Follow-up arms were added after
the nine-configuration evaluation, and the trace diagnosis is post-hoc. None of these analyses
is described as public preregistration or prospective clinical validation.

Finally, no clinicians evaluated downstream patient outcomes, and no workflow
was deployed in patient care. The observed success and unsafe-state patterns
therefore support an engineering conclusion: clinical agents should be tested
against both answer and state contracts, with recovery mechanisms evaluated
under explicit evidence and consistency assumptions. They do not establish
that \mg{} is ready to authorize clinical actions without human oversight.

\section{Methods}

\subsection{Clinical task substrates and partitioning}

The fault-recovery evaluation uses public MedAgentBench~\citep{jiang2025}
tasks and its official preloaded HAPI/FHIR environment. Deduplication is performed
before the patient-group development/test split. The resulting 300-task
inventory contains 30 tasks for each of ten templates; 67 tasks require a
write, of which 60 are in the 255-task test partition. Our benchmark layer
adds explicit answer/state contracts and eligible tool disturbances. We do
not use or relabel this substrate as MedAgentBench-v3.

Task expansion contains 565 total variants, with 470 in the test partition.
It changes numeric/temporal decision conditions, adds preexisting records,
and constructs explicit operational actions over external-source data.
Branches retain parent-task and patient-group identifiers for analysis.
The decision-branch arm has 76 parent tasks from 57 groups, and the external-action
arm has 69 parents from 69 groups with 138 test variants. These constructions
are not newly collected clinical labels or validated therapeutic policies.

External read-only questions come from FHIR-AgentBench~\citep{lee2026} and
MIMIC-IV Clinical Database Demo on FHIR version 2.1.0~\citep{bennett2025},
which builds on the MIMIC-IV-to-FHIR~\citep{bennett2023} representation
and is distributed through PhysioNet~\citep{pollard2026}.
The external evaluation selects 100 questions, 45 patients and 25 templates
from patients not attempted in an earlier, unfinished external evaluation.
The development set
contains eight questions from four patients. The selected test questions are
identical across the four models. Source reference answers are retained.
All stages use existing public data, not newly recruited participants.

\subsection{Failure injection and eligible assignments}

The six classes are interface drift, transient service failure, stale or
incomplete evidence, patient mismatch, provenance conflict and ambiguous
write outcome. The 806 assignments per policy in the nine-configuration evaluation comprise 152 interface
drift, 255 transient-503, 71 stale, 104 patient-mismatch, 104 provenance-conflict,
60 committed-timeout and 60 uncommitted-timeout assignments. The last two
are variants of one class, not a seventh conceptual class.

Interface drift rejects the \texttt{patient} observation-search parameter
while public metadata exposes \texttt{subject}. A transient failure returns
a service error at the eligible operation. Read-side disturbances supply a
stale observation, distractor-patient resources, or a conflicting preliminary
cache observation, respectively. The authoritative read remains documented.
For a committed timeout, the underlying write executes before the proxy
returns an outcome-unknown response; for an uncommitted timeout, the same
response is returned without the write. Eligibility is determined from the
task construction rather than selected after observing the model's response.
Assigned faults remain in the denominator even if the agent never reaches
their trigger. Trigger counts are recorded separately.

The external read-only adapter uses interface drift, transient-503, partial
history, patient mismatch and provenance conflict. Partial history is an
external analogue of missing/stale read evidence, not an identical perturbation
to the MedAgentBench-derived stale-value intervention. It does not inject write failures.

\subsection{MedGuard and policy controls}

\mg{} wraps public tool calls. For targeted observation reads it obtains the
authoritative response, follows trusted pagination, verifies patient and code
references, retains final/amended/corrected observations and preserves requested
ordering and count semantics. It can recover documented interface drift through
metadata and retry transient read failures within a bounded loop. It does not
discover all clinical evidence sources or infer arbitrary temporal requirements
that were absent from the request. If authoritative verification fails, the
evaluated implementation returns an explicit generic verification error;
this behavior is retained in the adverse trace analysis.

Before a patient-linked POST, the wrapper queries equivalent existing records
using the benchmark's normalized clinical resource representation. After an
ambiguous server response, it repeats the query and compares record identities.
The comparison assumes a single writer and immediate read visibility. Exactly
one new equivalent record is recognized as a commit without reissuing the
write; if no new equivalent record is found, the wrapper retries the write
once. Failed verification or multiple new candidates leave the commit
unestablished. This logic is not a transaction-level exactly-once guarantee.
The guard does not access hidden task contracts or gold outputs; all baseline
agents have access to the same documented public recovery interfaces.

The nine-configuration evaluation compares vanilla, generic warning, ordinary retry, a fault-class
hint and \mg{}. The hint names the disturbance class, not the reference answer
or an optimal policy. In the matched-control experiment, informed execution adds an explicit checklist for
identity, evidence and write-outcome verification. Evidence-only execution
attaches raw authoritative or pre/post record evidence without filtering it,
declaring commit or automatically repeating the write. Component deletions
remove read verification, write attribution, or generic retry/interface repair.
The last deletion retains verification and committed-write recognition; it
must not be interpreted as removing every possible recovery pathway.

\subsection{Models, inference and protocol adaptations}

The nine-configuration evaluation uses Qwen3.5-9B~\citep{qwen352026},
Ministral-3-8B-Instruct-2512~\citep{liu2026ministral},
Qwen2.5-7B-Instruct~\citep{qwen252024},
Qwen2.5-14B-Instruct~\citep{qwen252024}, Qwen3-8B~\citep{yang2025qwen3},
Llama-3.1-8B-Instruct~\citep{grattafiori2024}, Gemma-3-12B-it~\citep{gemma2025},
and Phi-4-mini-instruct~\citep{abouelenin2025}. Qwen3 contributes thinking and
non-thinking configurations. Follow-ups use Qwen3.5, Ministral, Llama and
GPT-OSS-20B~\citep{openai2025gptoss};
the targeted adverse-action analysis also includes Phi.

The nine-configuration evaluation uses a ceiling of 12 model turns, 2,048 new tokens per turn
and a 16,384-token context. The mechanism, task-expansion
and call-budget experiments use 12 turns,
2,048 output tokens, 16,384 context tokens and a 48-raw-call ceiling,
except the declared sensitivity arms. Greedy decoding is standard;
sampling experiments use temperature 0.7.
All raw public operations, including guard helper operations, count toward
the follow-up call ceiling.

External question answering uses 20 turns, 2,048 output tokens per turn, a 16,384-token context,
48 raw calls and batch size two.

\subsection{Mechanism, stress and budget experiments}

Repeated faults allow three eligible occurrences; compound faults combine
eligible components while tracking their triggers separately. An operation-matched
transient arm places the error on the agent's target operation so guard preflight
cannot consume it. Delayed visibility masks a committed record from the next
relevant verification read. The concurrency arm inserts a controlled equivalent
external write and records its identity separately. These boundary violations
are not declared recoverable cases and remain in the denominators.

Token-ceiling and sampling experiments use a fixed 50-task subset. The shared-call-ceiling experiment uses
50 tasks from 40 groups and four policies, with 50 clean and 160 fault
assignments per policy. Its 12-call ceiling is reachable within the 12-turn
limit for every policy; a larger advertised allowance that vanilla cannot
exhaust would not provide the same control. This is a shared maximum, not a
claim of matched realized computation.

\subsection{Grading and statistical analysis}

For task $t$, policy $p$ and eligible disturbance $h$, define
\begin{equation}
Y_{tph}=\mathbf{1}\{A_{tph}\land S_{tph}\land\neg U_{tph}\},
\end{equation}
where $A$ is the answer contract, $S$ is the required-state contract and $U$
is any forbidden mutation. Recorded violations include wrong-patient writes,
duplicates, unnecessary actions, incorrect-content creations and collateral
mutations. End-state and any transient unsafe states are distinguished in
the exported metrics. The normalizer is a benchmark profile, not a universal
test of clinical equivalence for all valid FHIR representations.

Let $H_t$ be the assigned non-clean disturbances for a task and let
$\mathcal C_m$ contain tasks solved by model configuration $m$ under
vanilla-clean execution. The primary conditional effect is
\begin{equation}
\widehat\Delta_{m,\mathcal C}=
\frac{1}{|\mathcal C_m|}\sum_{t\in\mathcal C_m}
\frac{1}{|H_t|}\sum_{h\in H_t}
\left(Y_{t,\mathrm{MG},h}-Y_{t,\mathrm{V},h}\right).
\end{equation}
The whole-task effect replaces $\mathcal C_m$ with all test tasks. Follow-up
effects average paired branches/faults within parent tasks before averaging
parents. Patient/group-cluster bootstrap resamples preserve correlated tasks;
10,000 draws yield 95\% percentile intervals. Pooled sampling results are averaged within tasks
before group resampling. No confirmatory $p$-values or multiplicity-adjusted
significance declarations are used.

For external question answering, $Y$ instead denotes original-answer correctness. Patients are
weighted equally, then their selected questions equally, then the five assigned
faults equally. Patient-cluster bootstrap uses 10,000 draws.
The primary external comparisons are guarded-minus-vanilla and
guarded-minus-informed over all questions. Clean-capable external analyses
are secondary. Infrastructure completeness is audited separately; final
external-evaluation coverage is 1,800/1,800 per model, with zero missing assignments.
Turn/context exhaustion counts as an observed model outcome. No sample size
was calculated from an assumed clinical effect; coverage is determined by the
available constructed tasks and computational protocol. Analysis is
descriptive and finite-benchmark, not population prevalence estimation.

\section*{Data availability}

Source tasks are available from MedAgentBench
(\url{https://github.com/stanfordmlgroup/MedAgentBench}) and FHIR-AgentBench
(\url{https://github.com/glee4810/FHIR-AgentBench}). External clinical records
are from MIMIC-IV Clinical Database Demo on FHIR, version 2.1.0
(\url{https://physionet.org/content/mimic-iv-fhir-demo/2.1.0/}). Aggregated
Source Data and task-selection indices are available at
\url{https://github.com/Learner4everrr/MedToolFault-MedGuard}.

\section*{Code availability}

The MedToolFault and MedGuard code is available at
\url{https://github.com/Learner4everrr/MedToolFault-MedGuard}.

\section*{Ethics statement}

This study analyzes publicly available benchmark tasks and demo data from
MedAgentBench, FHIR-AgentBench and MIMIC-IV Clinical Database Demo on FHIR.
All evaluations were performed in computational environments using existing
data. The research involved no participant recruitment, patient intervention
or deployment in clinical care.

\section*{Acknowledgements}

This work was supported by the National Institutes of Health's National Center
for Complementary and Integrative Health under grant numbers R01AT009457 and
U01AT012871, the National Institute on Aging under grant number R01AG078154,
the National Cancer Institute under grant number R01CA287413, the National
Institute of Diabetes and Digestive and Kidney Diseases under grant number
R01DK115629, and the National Institute on Minority Health and Health
Disparities under grant number 1R21MD019134-01. Zaifu Zhan was also supported
by the University of Minnesota Doctoral Dissertation Fellowship.

\section*{Author contributions}
Z.Z. conceived and designed the study, developed MedToolFault and MedGuard,
and implemented the software. Z.Z. conducted the experiments, analyzed the
results, and drafted and revised the manuscript. R.Z. supervised the research
and provided scientific guidance throughout the project. Z.Z., M.Z., S.Z.,
X.C., K.Y., M.S., Y.W., Y.R., S.C., M.C. and R.Z. contributed to scientific
discussions on the study design, experimental findings and interpretation of
the results, and approved the final version of the manuscript.

\section*{Competing interests}
The authors declare no competing interests.

\input{references.tex}
\end{document}

%% file: authors.tex
\author{\parbox{0.95\textwidth}{\centering
Zaifu~Zhan\textsuperscript{1}, Min~Zeng\textsuperscript{1},
Sicheng~Zhou\textsuperscript{2}, Xiaoyi~Chen\textsuperscript{1},
Kai~Yu\textsuperscript{1}, Meijia~Song\textsuperscript{1},
Yifan~Wu\textsuperscript{1}, Yang~Ruan\textsuperscript{1},
Saeid~Cheshmi\textsuperscript{1}, Mengyuan~Cui\textsuperscript{1},
Rui~Zhang\textsuperscript{1,*}\\[5pt]
{\normalsize\mdseries
\textsuperscript{1}University of Minnesota, Minneapolis, MN, USA\\[2pt]
\textsuperscript{2}GE HealthCare, Bellevue, WA, USA\\[4pt]}
{\small\mdseries \textsuperscript{*}Correspondence: Rui Zhang (\href{mailto:ruizhang@umn.edu}{\texttt{ruizhang@umn.edu}}).}
}}

%% file: figure_pages.tex
\newcommand{\OverviewFigure}{%
\begin{figure}[!htbp]
\centering\includegraphics[width=\linewidth,height=.77\textheight,keepaspectratio]{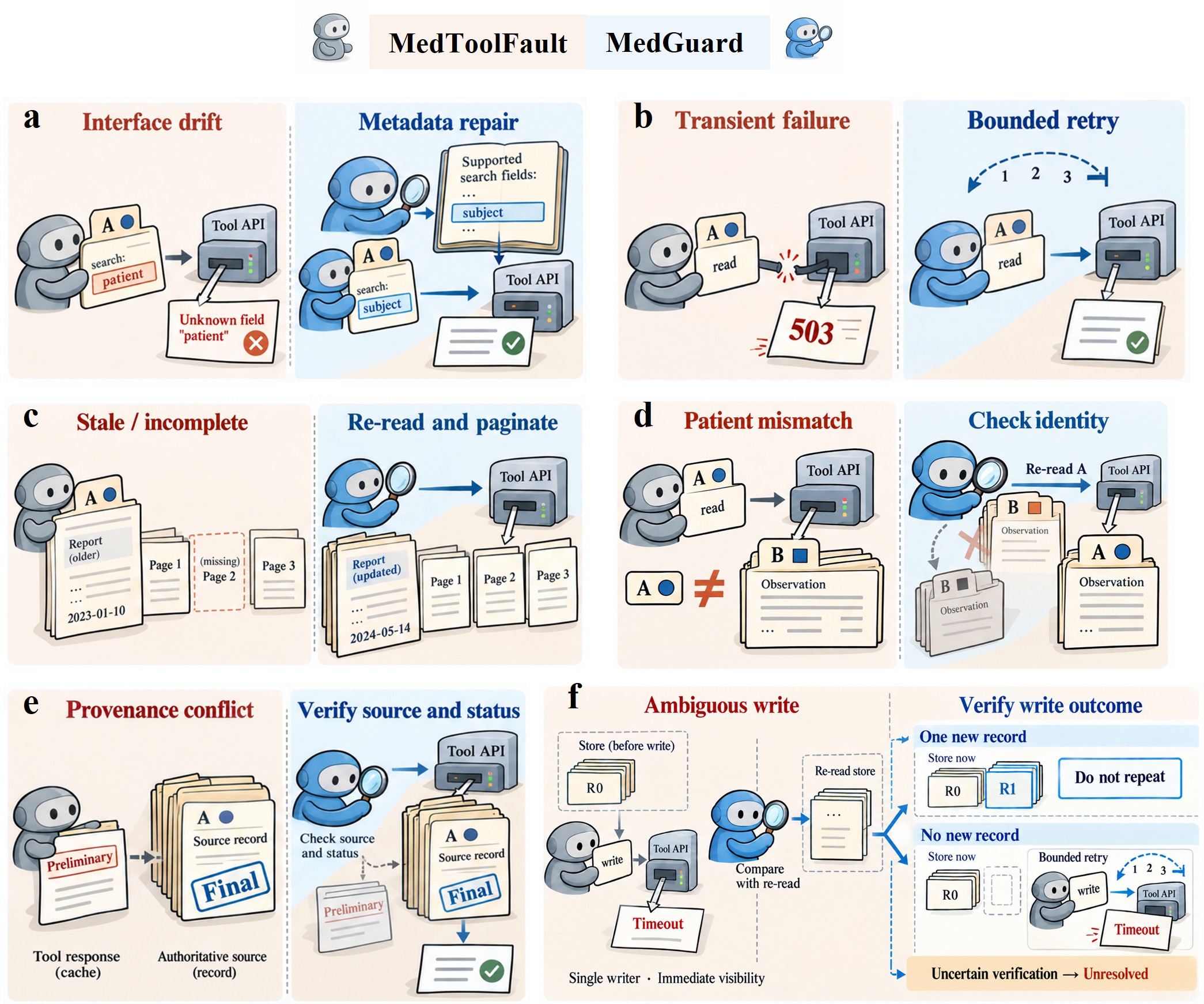}
\caption{\textbf{Tool failures introduced by MedToolFault and recovery operations performed by MedGuard.}
Each scene pairs a tool failure with its public-interface recovery mechanism.
\textbf{(a)} Interface drift: consult public metadata and repair an unsupported
request parameter. \textbf{(b)} Transient failure: retry a failed read within a
bounded budget. \textbf{(c)} Stale or incomplete evidence: re-read authoritative
observations and follow pagination to recover query-relevant history. Stale
observations and incomplete external histories are distinct interventions.
\textbf{(d)} Patient mismatch: check patient references and re-read the requested
patient's record, without relabelling another patient's evidence.
\textbf{(e)} Provenance conflict: verify source and finalized status against
authoritative records. \textbf{(f)} Ambiguous write: compare equivalent records
before and after a write whose response times out.}
\label{fig:overview}
\end{figure}
}

\newcommand{\RecoveryFigure}{%
\begin{figure}[!htbp]
\centering\includegraphics[width=\linewidth,height=.77\textheight,keepaspectratio]{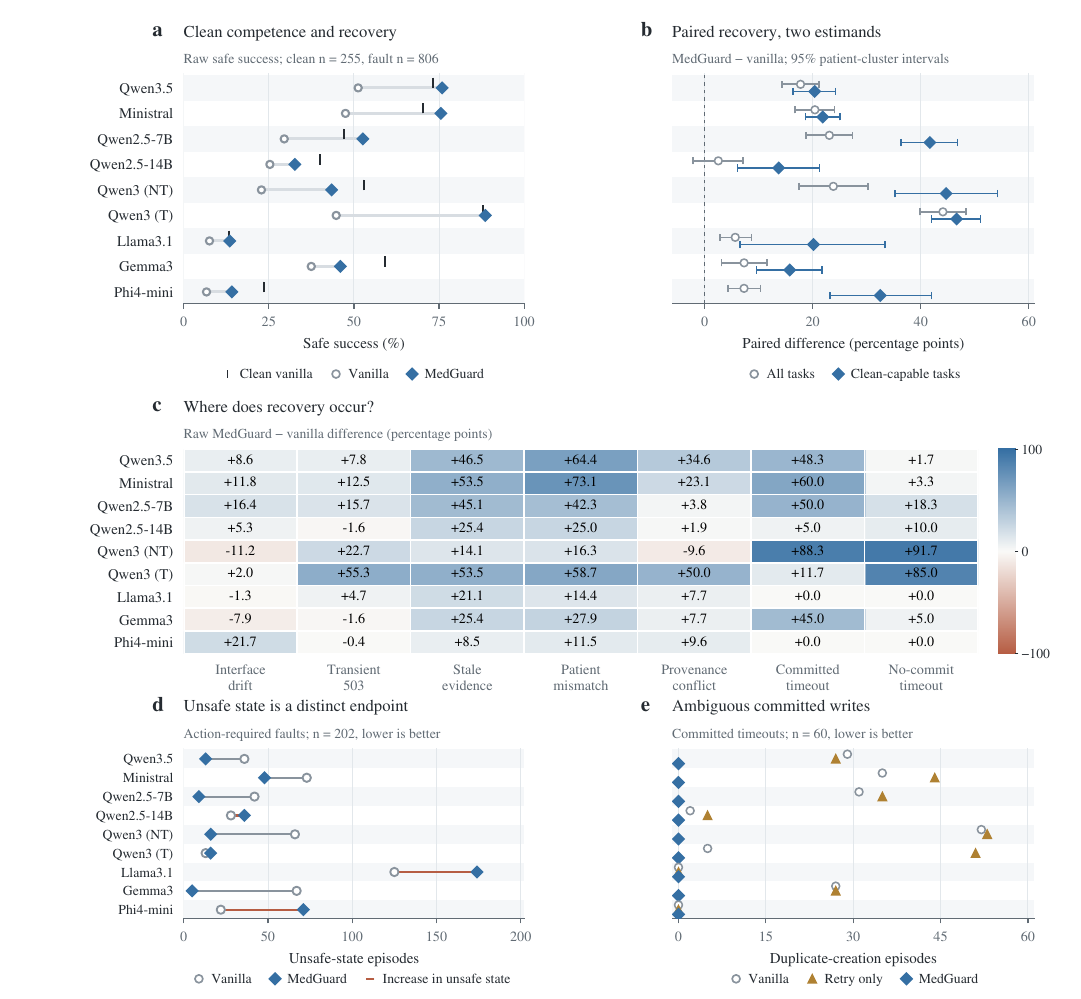}
\caption{\textbf{MedGuard recovery and clinical state endpoints on MedToolFault.}
\textbf{(a)} Raw safe-success percentages over 255 clean tasks or 806 assigned
faults per policy/configuration. Vertical ticks denote vanilla clean performance;
open circles and diamonds denote vanilla and MedGuard fault performance.
Connecting lines are visual guides, not uncertainty intervals.
\textbf{(b)} MedGuard-minus-vanilla paired effects
for all tasks and each configuration's vanilla-clean-capable subset. Points are
task-equal effects; bars are descriptive 95\% patient-cluster percentile
intervals from 10,000 draws. All-task analyses contain 88 groups; conditional
group counts and eligibility are included in Source Data. \textbf{(c)} Raw
per-hazard percentage-point differences, without paired weighting or uncertainty
claims. Fault denominators are 152, 255, 71, 104, 104, 60 and 60 from left to
right. \textbf{(d)} Unsafe-state counts among 202 action-required fault assignments.
Warm connecting lines flag increases, not statistically tested differences.
\textbf{(e)} Duplicate-creation counts among 60 committed-timeout assignments.
Zero duplicates do not imply successful completion or absence of other
violations. V, vanilla; MG, MedGuard; NT/T, non-thinking/thinking. Configurations
and task assignments are correlated, not independent clinical samples.
Source Data: raw endpoints, hazard rates and paired effects.}
\label{fig:primary}
\end{figure}
}

\newcommand{\MechanismsFigure}{%
\begin{figure}[!htbp]
\centering\includegraphics[width=\linewidth,height=.75\textheight,keepaspectratio]{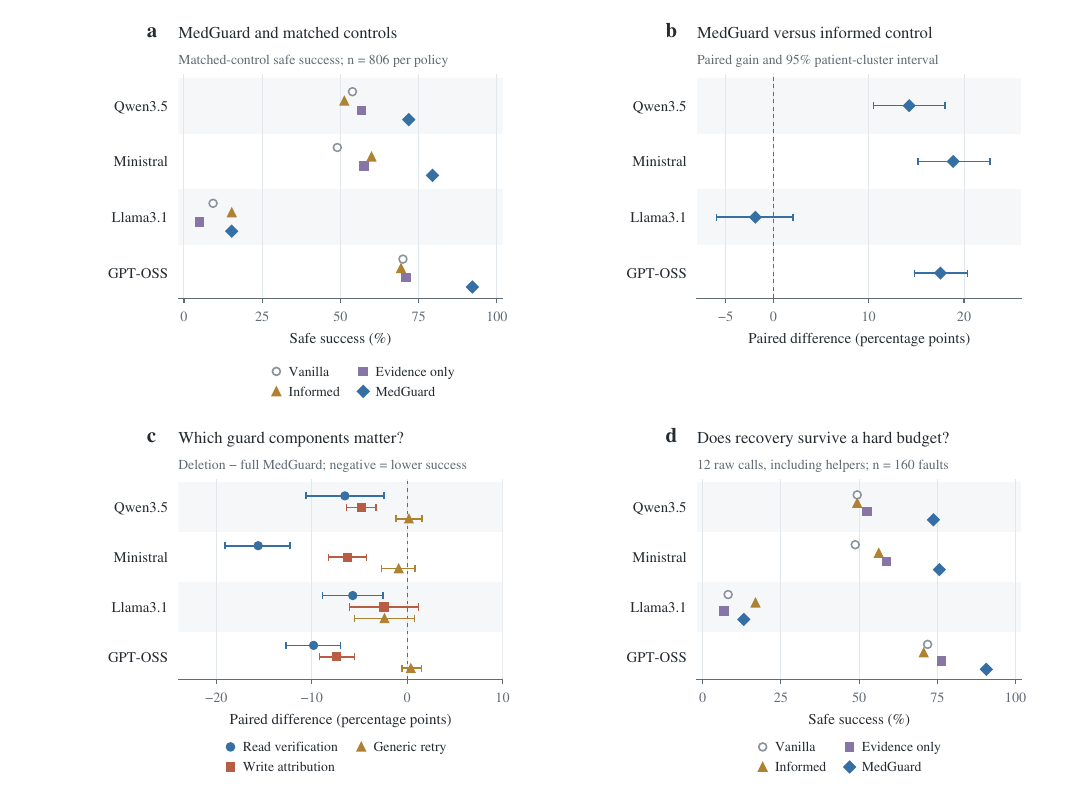}
\caption{\textbf{MedGuard controls, component ablations and a shared raw-call ceiling.}
\textbf{(a)} Matched-control fault safe success (806 assignments per policy/model;
255 parents, 88 patient groups). Evidence denotes evidence-only execution.
\textbf{(b)} Guarded-minus-informed paired effects.
\textbf{(c)} Component-deletion-minus-full-guard effects; read, write and retry
denote removal of read verification, write attribution and generic retry/interface
repair, respectively. The retry deletion retains other verification pathways.
\textbf{(d)} Safe success under a shared 12-raw-call ceiling (160 fault assignments,
50 tasks, 40 groups per policy/model); every helper call counts. Points and
bars in \textbf{(b), (c)} show parent-equal effects and 95\% patient-cluster intervals
from 10,000 draws. Raw rates in \textbf{(a), (d)} are not paired effects. No
multiplicity-adjusted claims are made. Source Data and Supplementary Tables~S3,
S4 and S6 provide counts and exact intervals.}
\label{fig:mechanisms}
\end{figure}
}

\newcommand{\StressFigure}{%
\begin{figure}[!htbp]
\centering\includegraphics[width=\linewidth,height=.76\textheight,keepaspectratio]{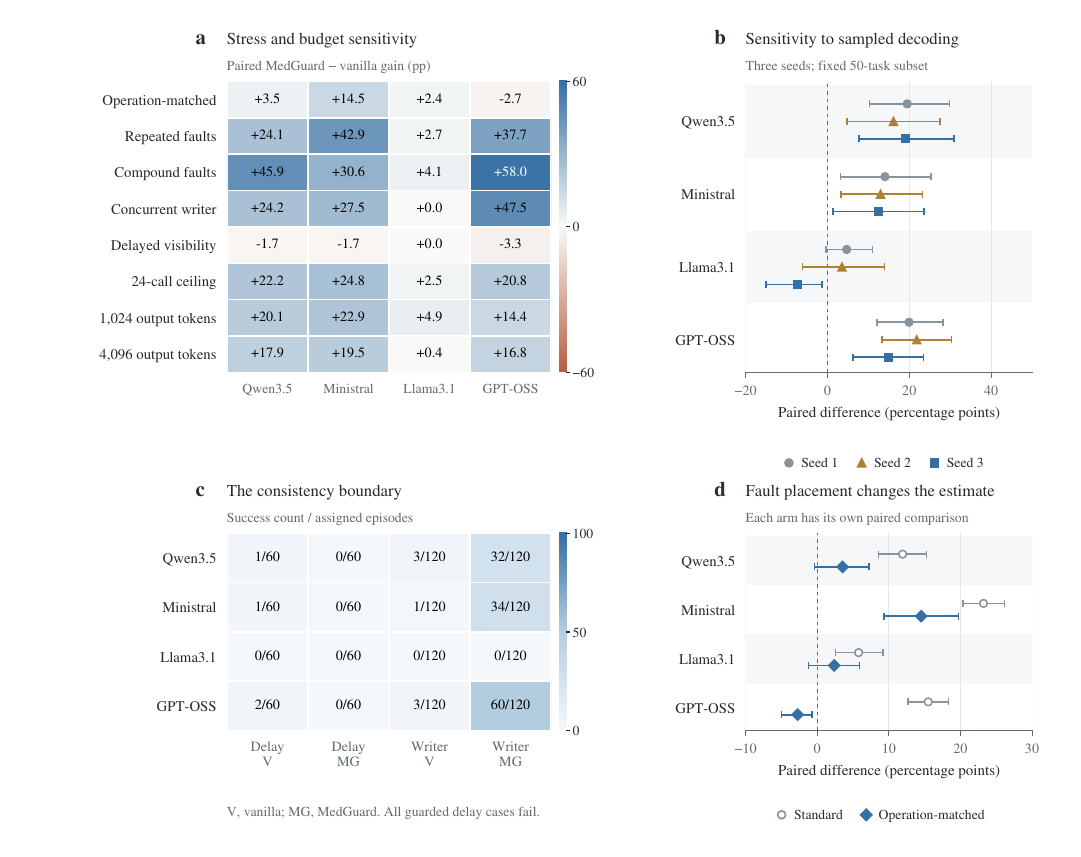}
\caption{\textbf{Stress sensitivity and unrecovered consistency boundaries.}
\textbf{(a)} Parent-equal guarded-minus-vanilla effects in eight stress/budget
settings. Color represents the point estimate only; exact intervals and
denominators are in Supplementary Table~S3 and Source Data.
\textbf{(b)} Sampling runs (temperature 0.7; fixed 50-task subset), shown
separately with 95\% patient-cluster intervals.
\textbf{(c)} Raw success counts under delayed visibility and a controlled
concurrent writer; background shading encodes the success percentage, while
text gives numerator/denominator. All four guarded models fail all 60 delayed
visibility assignments. Concurrent-writer state violations can include the
external writer's contribution.
\textbf{(d)} Standard matched-control and operation-matched transient effects.
Operation matching prevents guard preflight from consuming the target fault.
These are distinct arms, not a paired test of their difference. Intervals use
10,000 patient-cluster bootstrap draws. Repeated tasks and sampling runs are not
independent patient cohorts.}
\label{fig:stress}
\end{figure}
}

\newcommand{\TransferFigure}{%
\begin{figure}[!htbp]
\centering\includegraphics[width=\linewidth,height=.74\textheight,keepaspectratio]{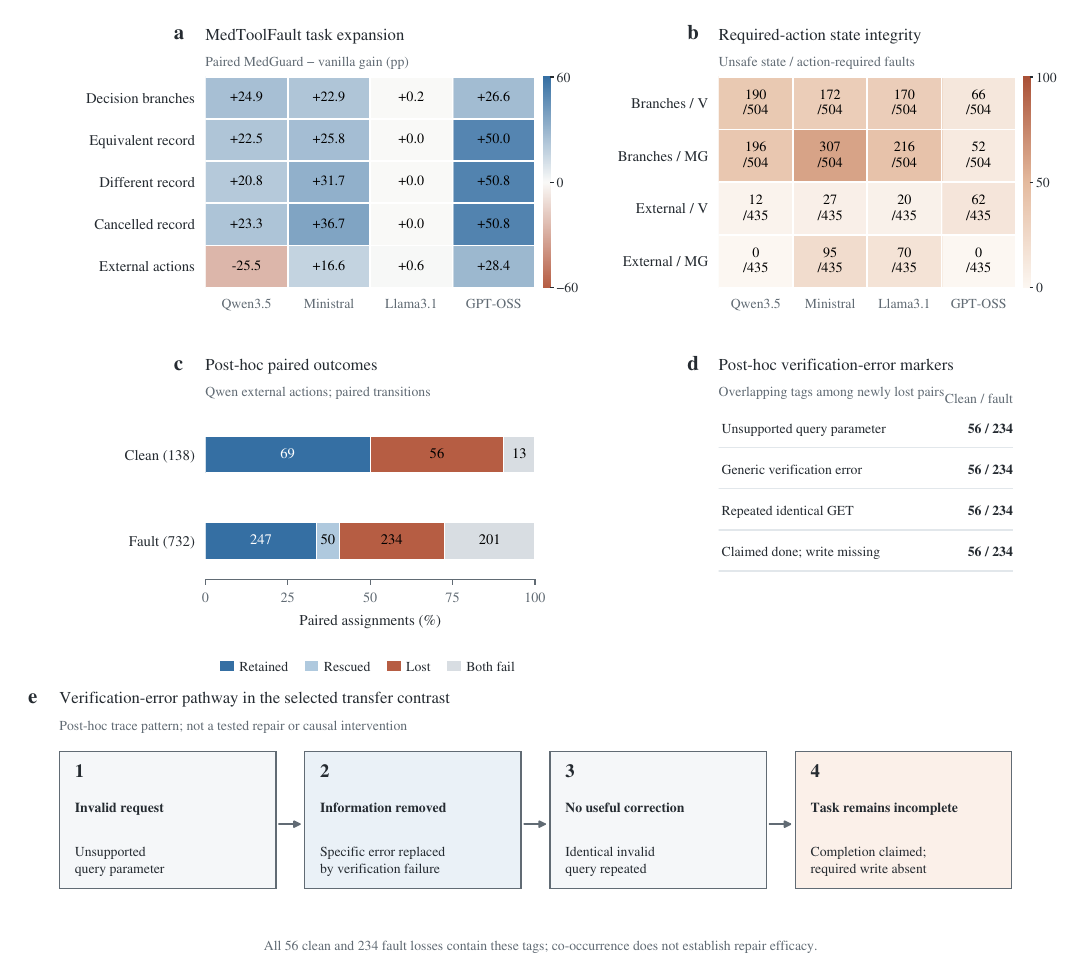}
\caption{\textbf{MedToolFault task expansion and post-hoc verification diagnostics.}
\textbf{(a)} Parent-equal paired gains across five expansion arms; exact intervals
are provided in Supplementary Table~S5. These are semi-synthetic operational
tasks, not treatment-effect evaluations. \textbf{(b)} Unsafe counts among
action-required faults; cell text supplies each denominator and background
encodes the percentage. \textbf{(c)} Qwen external-action paired transitions:
retained success, rescued failure, newly lost success, or failure under both
policies. Counts are printed within bars where space permits; zero is omitted.
\textbf{(d)} Co-occurring tags among all 56 newly lost clean and 234 newly lost
fault pairs. The right-hand entries are clean/fault counts, not fractions.
Tags overlap and are not mutually exclusive causes.
\textbf{(e)} An information-loss pathway consistent with the code and observed
traces. The analysis is post-hoc; no error-preserving replacement guard was
evaluated and no repair efficacy is inferred. Source Data include tag counts,
pair references and compressed-episode hashes. Missing required writes are
failures, not automatically forbidden mutations.}
\label{fig:transfer}
\end{figure}
}

\newcommand{\ExternalFigure}{%
\begin{figure}[!htbp]
\centering\includegraphics[width=\linewidth,height=.75\textheight,keepaspectratio]{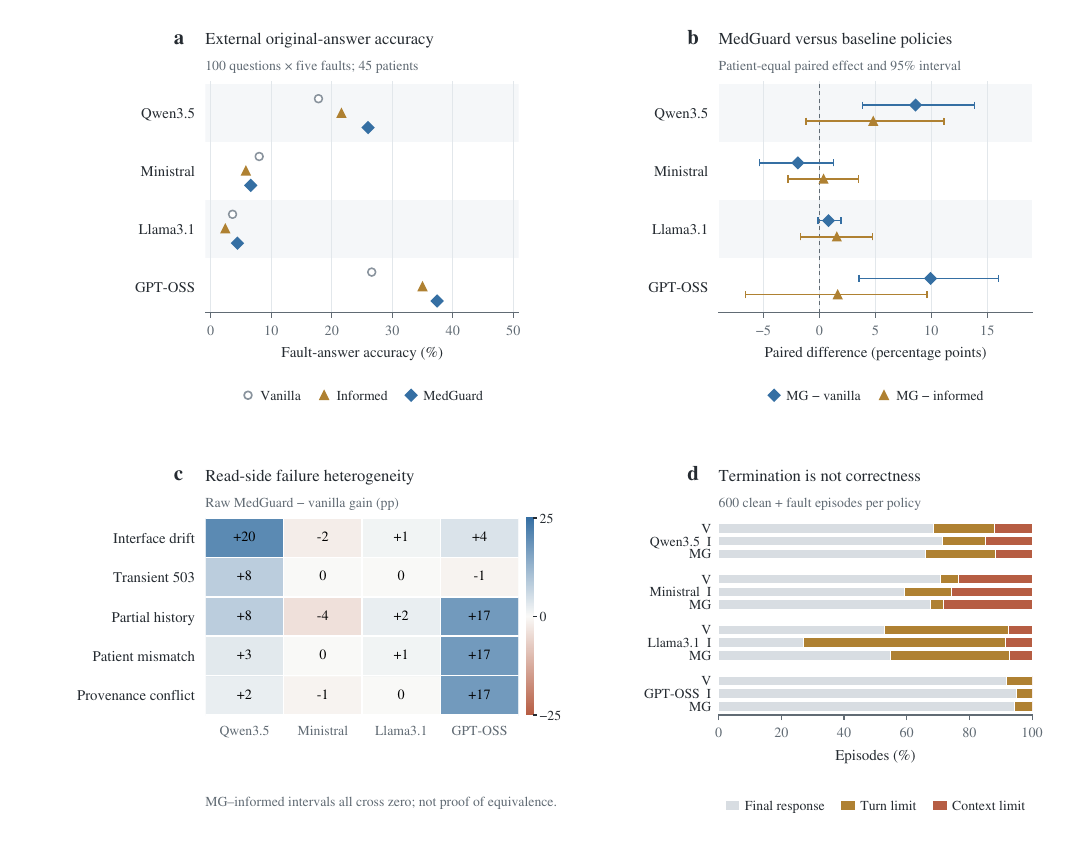}
\caption{\textbf{External original-answer question answering.}
\textbf{(a)} Raw accuracy over 500 fault assignments per policy/model (100
questions, five faults, 45 patients). \textbf{(b)} Patient-equal paired effects
against vanilla and informed baselines, with 95\% patient-cluster percentile
intervals from 10,000 draws. All informed-baseline intervals cross zero;
this is neither demonstrated superiority nor equivalence.
\textbf{(c)} Raw fault-specific differences over 100 questions per cell.
\textbf{(d)} Termination reasons across 600 clean-plus-fault episodes per
policy/model. Final termination does not imply a correct answer; context and
turn limits are observed outcomes, not missing infrastructure records.
All models use the same selected questions but disclosed model-specific
serialization adapters. This read-only test does not assess write safety,
clinical outcomes or independent-hospital generalization. Exact counts and
intervals are in Supplementary Tables~S7--S8 and Source Data.}
\label{fig:external}
\end{figure}
}

\newcommand{\ResourcesFigure}{%
\begin{figure}[!htbp]
\centering\includegraphics[width=\linewidth,height=.76\textheight,keepaspectratio]{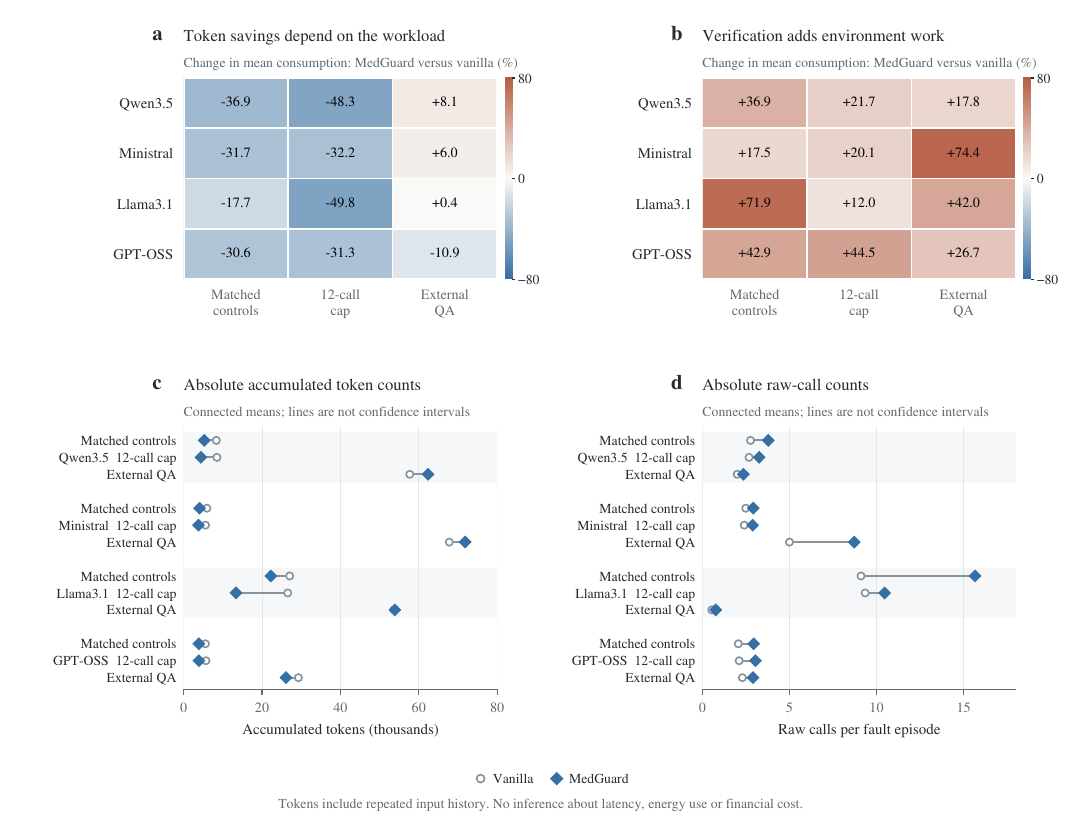}
\caption{\textbf{Accumulated token use and raw tool calls by workload and policy.}
\textbf{(a), (b)} Percentage changes in mean accumulated tokens and raw calls under
MedGuard relative to vanilla within each model and arm. Negative values mean
less consumption, not a loss of accuracy. Color is a signed change, not a
quality score. \textbf{(c), (d)} Absolute means with endpoints joined for each
paired workload; these lines are not confidence intervals. Workloads are matched
controls (806 fault assignments per policy), the shared 12-call ceiling
(160), and external read-only question answering (500). Tokens sum model input and
output over the episode, counting repeated history. Raw calls include automatic
helper operations. No between-model runtime normalization, latency estimate,
energy estimate or monetary saving is claimed. Complete means for vanilla,
informed and MedGuard are in Supplementary Table~S9 and Source Data.}
\label{fig:resources}
\end{figure}
}

%% file: tables/primary.tex
Qwen3.5-9B & 73.3 & 51.2 & 75.9 & 187 & $+20.37$ & $[16.40, 24.20]$ \\
Ministral-3-8B & 70.2 & 47.5 & 75.6 & 179 & $+21.86$ & $[18.67, 25.03]$ \\
Qwen2.5-7B & 47.1 & 29.5 & 52.6 & 120 & $+41.68$ & $[36.36, 46.77]$ \\
Qwen2.5-14B & 40.0 & 25.3 & 32.6 & 102 & $+13.71$ & $[6.07, 21.23]$ \\
Qwen3-8B (non-thinking) & 52.9 & 22.8 & 43.4 & 135 & $+44.69$ & $[35.18, 54.21]$ \\
Qwen3-8B (thinking) & 87.8 & 44.8 & 88.6 & 224 & $+46.64$ & $[42.00, 51.02]$ \\
Llama-3.1-8B & 13.3 & 7.6 & 13.5 & 34 & $+20.15$ & $[6.57, 33.33]$ \\
Gemma-3-12B & 59.2 & 37.5 & 46.0 & 151 & $+15.75$ & $[9.58, 21.77]$ \\
Phi-4-mini & 23.5 & 6.7 & 14.1 & 60 & $+32.50$ & $[23.17, 42.02]$ \\